\documentclass{article}
\usepackage{ijcai23}

\usepackage{times}
\usepackage{soul}
\usepackage{url}
\usepackage[hidelinks]{hyperref}
\usepackage[utf8]{inputenc}
\usepackage[small]{caption}
\usepackage{graphicx}
\usepackage{amsmath}
\usepackage{amsthm}
\usepackage{booktabs}
\usepackage{algorithm}
\usepackage{algorithmic}
\usepackage[switch]{lineno}
\usepackage{color}
\usepackage{float}
\usepackage{verbatim} %comments
\usepackage{apalike}
\usepackage{amssymb}
\usepackage{amsfonts}
\usepackage{threeparttable}
\usepackage{multirow}
\usepackage{graphicx}
\usepackage{subfig}
\usepackage{epsfig} 
\usepackage{epstopdf}
\title{DASTSiam: Spatio-Temporal Fusion and Discriminative Augmentation for Improved Siamese Tracking}

\author{
    Author Name
    \affiliations
    Affiliation
    \emails
    email@example.com
}

\author{
Yucheng Huang$^{1*}$
\and
Eksan Firkat$^{1*}$
\and
Ziwang Xiao$^{1}$
\and
Jihong Zhu$^{1,2}$
\and
Askar Hamdulla$^1$
\thanks{Yucheng Huang and Eksan Firkat make an equal contribution.}
\thanks{Askar Hamdulla is the corresponding author.}
\affiliations
$^1$School of Information Science and Engineering, Xinjiang University, Urumqi, China \\
$^2$Department of Precision Instrument,Tsinghua University, Beijing, China
\emails
\{107552101321, eksan, 107552103759\}@stu.xju.edu.cn.com,
jhzhu@tsinghua.edu.cn,
askar@xju.edu.cn
}

\begin{document}

\maketitle

\begin{abstract}
 Tracking tasks based on deep neural networks have greatly improved with the emergence of Siamese trackers. However, the appearance of targets often changes during tracking, which can reduce the robustness of the tracker when facing challenges such as aspect ratio change, occlusion, and scale variation. In addition, cluttered backgrounds can lead to multiple high response points in the response map, leading to incorrect target positioning. In this paper, we introduce two transformer-based modules to improve Siamese tracking called \textbf{DASTSiam}: the spatio-temporal (ST) fusion module and the Discriminative Augmentation (DA) module. The ST module uses cross-attention based accumulation of historical cues to improve robustness against object appearance changes, while the DA module associates semantic information between the template and search region to improve target discrimination. Moreover, Modifying the label assignment of anchors also improves the reliability of the object location. Our modules can be used with all Siamese trackers and show improved performance on several public datasets through comparative and ablation experiments. The code available at {\it https://github.com/huangliqwe2020/DASTASiam.}
\end{abstract}

\section{Introduction}
% introductio1 part 介绍
Object tracking is a fundamental problem in computer vision, with numerous applications in areas such as unmanned aerial vehicles, precision guidance, mobile robotics, and video surveillance. One of the most widely used and promising approaches to object tracking is the Siamese-based tracking algorithm~\cite{bertinetto2016fully,li2018high,guo2017learning,zhang2018structured,zhang2019deeper,wang2019fast,li2019siamrpn++}. These algorithms employ a two-branch neural network to track the target by inputting a template image representing the target and a search area image containing the target to a depth convolutional neural network-based backbone for feature extraction. A correlation function is then applied to the extracted features, generating a response map that indicates the target's location in the search area. Recent research has focused on improving the performance of tracking algorithms by Siamese network. For instance, SiamFC~\cite{bertinetto2016fully} refers to DCF trackers~\cite{henriques2014high,danelljan2017eco,valmadre2017end,danelljan2015learning,li2018learning} and uses Siamese network to train correlated filter, and SiamRPN~\cite{li2018high} refers to Fast-RCNN\cite{girshick2015fast} and proposes regional proposal network-based anchor settings. These methods offers a good balance between accuracy and real-time performance.
\begin{figure}[t!]
\centering
\subfloat{
    \includegraphics[width=0.8\linewidth]{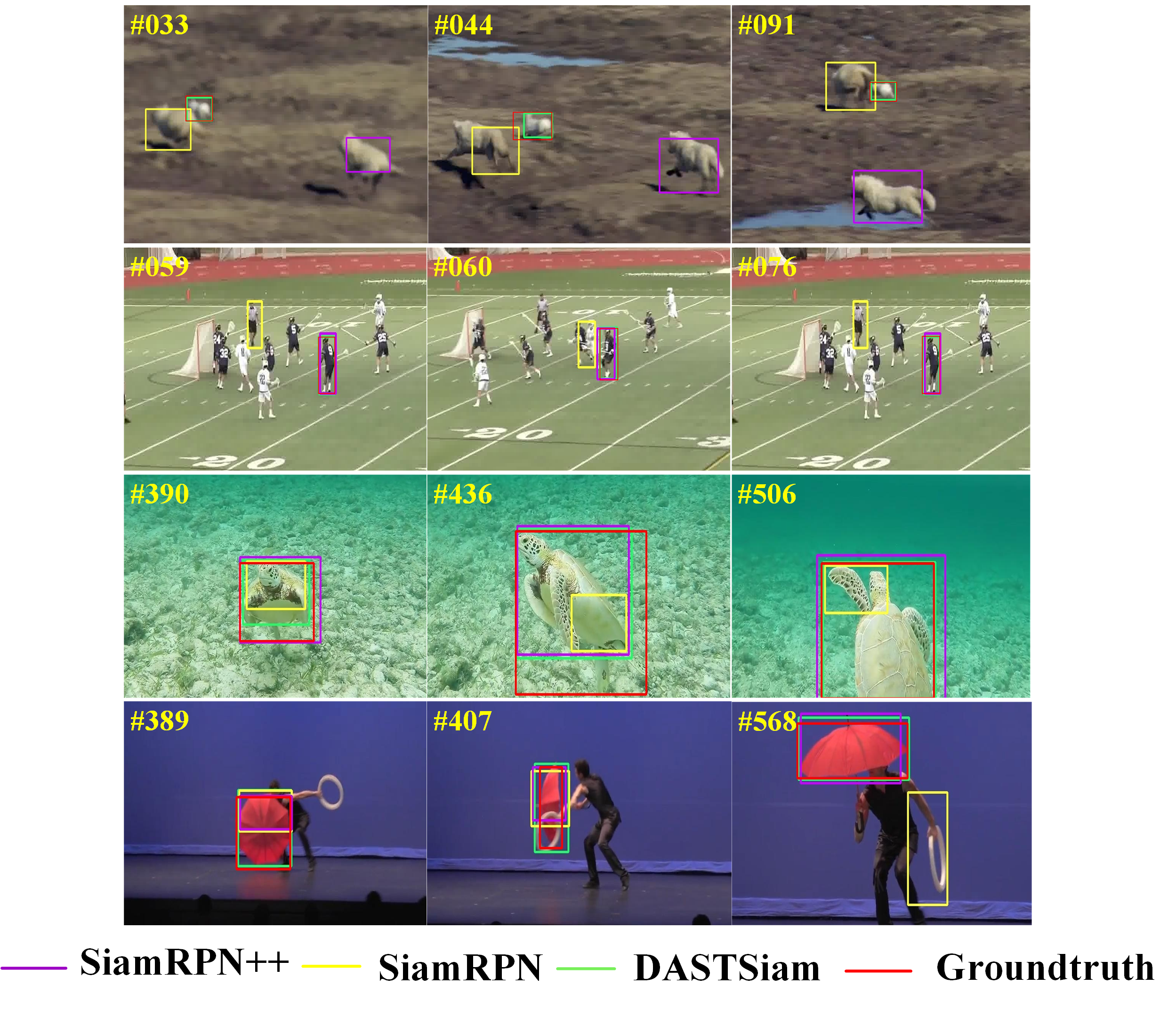}}
\caption{\textbf{Qualitative comparison.} With the help of Spatio-temporal module(ST) and discriminative augmentation(DA), Our \textbf{DASTSiam} can cope with appearance changes, motion blur and background clutter. Furthermore, it shows a better performance than SiamRPN and SiamRPN++.}  
\label{fig1} 
\end{figure}

Although the above Siamese trackers have obtained outstanding tracking performance, they did not exploit the Spatio-temporal information enough. In SiamFC and SiamRPN, the initial template is not updated, which can lead to tracking failure when appearance of targets changes during tracking. To address this issue, more recent Siamese trackers have implemented a linear updating technique\cite{li2018high,zhu2018distractor} to modify the template based on the appearance in the new frame. However, in complex environments with factors such as lighting changes and occlusion, it is important to learn an adaptive target descriptor that can handle such variations. Cluttered backgrounds also a common challenge in Siamese trackers, as they can easily lead to multiple high response points in the final response map, causing the tracker to fail to accurately locate the target due to a lack of depth semantic information to distinguish the target from interference. Recent developments such as DaSiamRPN \cite{zhu2018distractor} and RASNet \cite{wang2018learning} have attempted to address this issue by constructing negative sample sets of semantic interference on training samples and continuously learning to assign more weight to channels depicting target semantic information, respectively. However, these methods come with the difficulty of extra sample construction and do not fully exploit the semantic differences between the target and interference in the depth feature space.

% we introduce two transformer-based modules to improve Siamese tracking: the spatio-temporal (ST) module and the Discriminative Augmentation (DA) module.

In order to solve these problems, inspired by UpdateNet\cite{zhang2019learning} and some visual tasks\cite{parmar2018image,carion2020end,zeng2020learning} based on transformer, we propose a new method for enhancing Siamese tracking by utilizing the self-attention mechanism of transformer\cite{vaswani2017attention} called \textbf{DASTSiam}. Our method addresses the limitations of existing Siamese trackers in terms of their ability to exploit spatio-temporal information and distinguish targets from cluttered backgrounds. To improve the robustness of the tracker against target appearance changes, we introduce a portable spatio-temporal (ST) module. This module bridges isolated video frames and conveys rich temporal information across them.  To improve the tracker's ability to distinguish targets in cluttered backgrounds, we propose a discriminative augmentation (DA) module. This module strengthen the relevance of internal semantic similarity features of the search region feature and cross-attention mechanisms to exploit the semantic similarity between the target template and the search region. We also include two adaptive filters in our proposed method to improve the reliability of the template update and the output of the modified decoder. 

Overall, our method improves the generalization of target template features, fully exploits temporal cues in the tracking process, and enhances the ability to distinguish between the target and background in the search area (as shown in Figure~\ref{fig1}). Our contributions are summarized as follows.

\begin{itemize}
  \item We introduce a portable spatio-temporal module (ST) to bridges isolated video frames and fully exploit the temporal cues in the tracking process. The robustness of the tracker to target deformation, occlusion, and other problems is enhanced.

  \item We introduce a discriminative augmentation module (DA) to enhance the ability to distinguish between target and background in the tracking process.
  
  \item Experiments on the public benchmark verify that the proposed \textbf{DASTSiam} achieves the new state-of-the-art performance on several public datasets.
\end{itemize}

\begin{figure*}[t!]
	\centering
	\subfloat{
		\includegraphics[width=0.8\textwidth,height=0.3\textwidth]{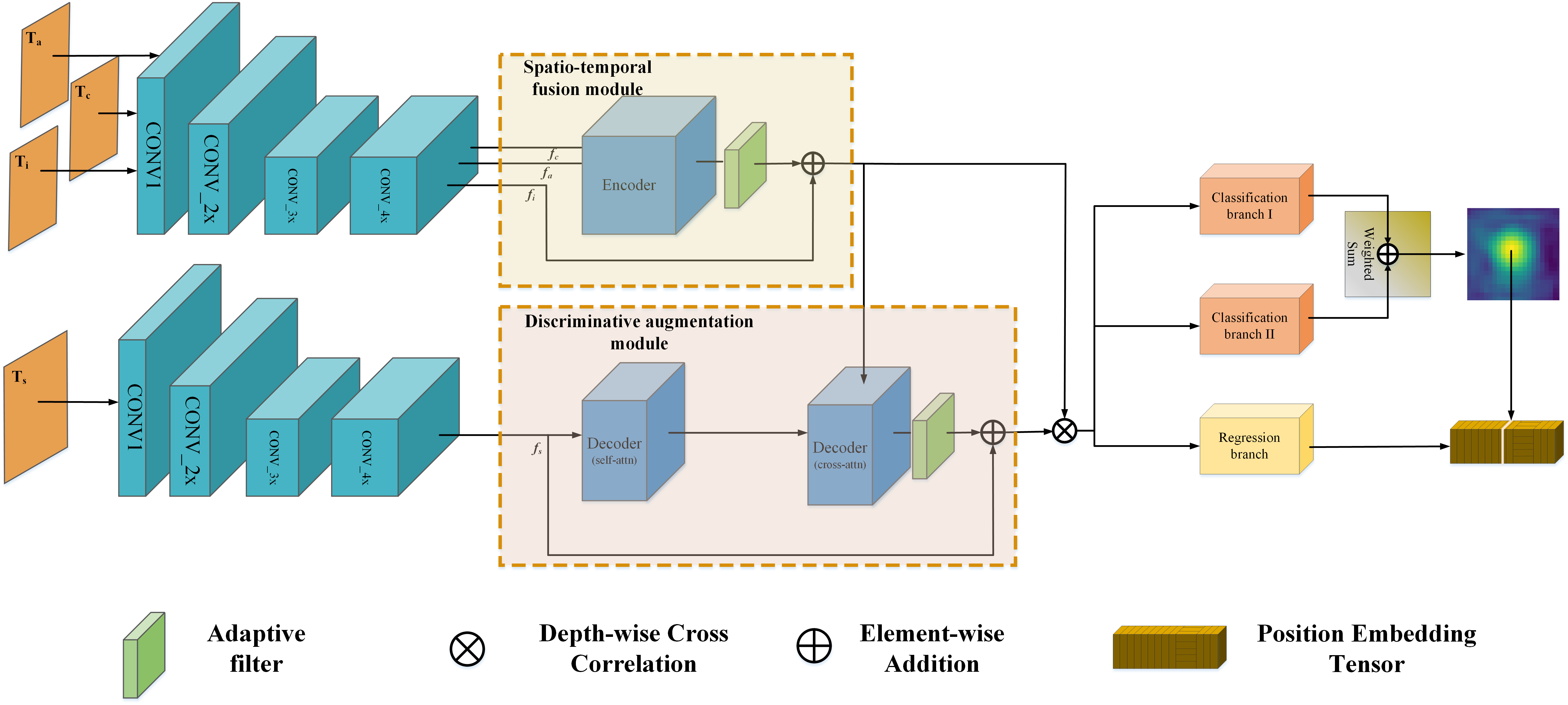}}
	\caption{\textbf{Overview of our tracking framework based on Siamese network with our proposed modules.}}
	\label{fig2} 
\end{figure*}

\section{Methods}

The proposed \textbf{DASTSiam} framework is detailed in this section. The framework includes two key components: the spatio-temporal (ST) fusion module  and the discriminative augmentation module (DA). The ST module improves the adaptability of template updating to handle changes in the target's appearance by modifying the encoder. The DA module enhances the discriminative ability of the tracker by creating a semantic association between the matching template and the search area in the depth feature space, using modifications in the decoder. A visual representation of the \textbf{DASTSiam} framework can be found in Figure~\ref{fig2}.

 \begin{figure}[t!]
\centering
\subfloat{
\includegraphics[width=0.3\textwidth,height=0.3\textwidth]{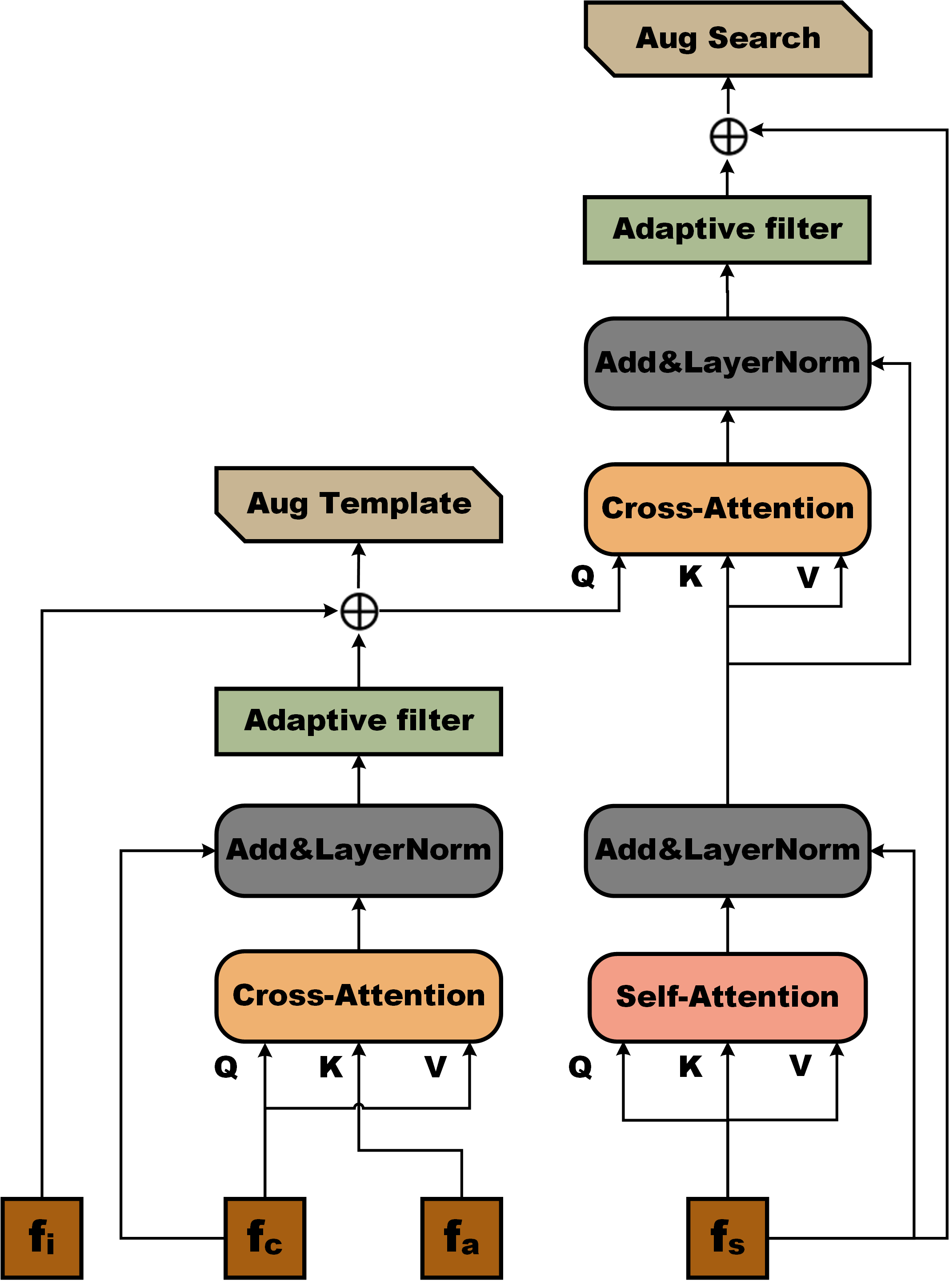}}
\caption{An overview of the proposed transformer based spatio-temporal (ST) module and discriminative augmentation module(DA).}
\label{fig3} 
\end{figure}
\subsection{Spatio-temporal (ST) fusion module.}
\quad  In the Siamese network, the process of extracting features from the template image $T_{z}$ and the search image $T_{x}$ is done by using a backbone network. These extracted features are represented by $f_{z}$ and $f_{x}$ respectively. Cross-correlation operation is performed between these features to generate a response map. To enable the tracking system to make use of temporal cues and inter-frame relations, we propose to use spatio-temporal fusion module (ST) on $f_{z}$ feature maps before cross-correlation operation. This approach is formulated as:
\begin{small}  
\begin{equation}
	ST(f_{i},f_{a},f_{c})=\Phi_{ST}(Encoder(f_{a},f_{c}))+f_{i}
\end{equation}
\end{small}  
Our proposed encoder uses an attention mechanism to expand the receptive field and obtain more context information, thus improving the utilization of spatial information in the spatio-temporal fusion module (ST). To efficiently exploit temporal cues and improve inference speed, we use a cross-attention mechanism for multi-frame fusion instead of channel-wise concatenation. We select three frames from the same video sequence, $T_{i}$, $T_{a}$, and $T_{c}$, and extract their features to obtain $f_{i}$, $f_{a}$, and $f_{c}$ respectively. Then, we use $f_{a}$ and $f_{c}$ as the input of the encoder, with $f_{c}$ as the query and value and $f_{a}$ as the key. The feature fusion is done through the cross-attention mechanism, which is mathematically represented as:
\begin{equation}
CrossAttn(Q,K,V)=Softmax(\dfrac{Q_{c}K^{T}_{a}}{\sqrt{d}})V_{c}\label{equation5}
\end{equation}
The cross-attention mechanism maps the permuted feature maps $f_{a}$ and $f_{c}$ through three different fully connected layers (FC) to obtain $Q_{c}$, $V_{c}$, and $K_{a}$, which are in the dimension of $\mathbb{R}^{\frac{N_{t}}{2stride}\times \frac{N_{t}}{2stride} \times C}$. $(\frac{N_{t}}{2stride})^{2}$ is the number of feature points in the feature map, and each point contains $C$ different feature information, enhancing the semantics of each feature point. The FC layers have a weight dimension of $(N_{c}, N_{c})$ where $N_{c}$ is the number of channels. As represented in Eq.(\ref{equation7}), each feature point $\hat{N}_{i}$ in $Q_{c}$, $V_{c}$, $K_{a}$ integrates all its channel-wise semantic information, which greatly enriches the spatial context of features.
\begin{equation}
\begin{aligned}
&\hat{N}_{i,j}=\dfrac{\sum_{j=0}^{N_{c}}w_{j}N_{i,j}}{N_{c}}\\
&N_{i}=(N_{i,0},N_{i,1},...,N_{i,j})_{N_{c}}\\
&\hat{N}_{i}=(\hat{N}_{i,0},\hat{N}_{i,1},...,\hat{N}_{i,j})_{N_{c}}
\end{aligned}
\label{equation7}
\end{equation}
$N_{i}$ refers to feature point in $f_{i}$,$f_{a}$,$f_{c}$.For convenience, mark each feature point in $Q_{c}$,$K_{a}$ as $\hat{Nc}_{i}$ and $\hat{Na}_{i}$ respectively. To enable ST to exploit temporal information, we directly use matrix multiplication to compute the attention matrix($AttnMat$) of $Q_{c}$ and $K_{a}$. Although we use FC with different weights for $Q_{c}$ and $K_{a}$, their mapping rules are the same, that is, the similarity between $Q_{c}$ and $K_{a}$ in the same spatial position can be measured by $AttnMat$. Most notably, $f_{a}$ contains rich and reliable historical information, which is the output of ST in the last stage. In this way, ${AttnMat}$ in Eq.(\ref{equation8}) can make $f_{c}$ pay more attention to those positions with highly Spatio-temporal similarity.
\begin{equation}
\begin{aligned}
&AttnMat_{i,j}=P_{i,j}(T_{c}|T_{i},T_{i+1},...,T_{c-1})\\
&\qquad\qquad\quad=\sum_{j=0}^{N_{c}}\hat{Nc}_{i,j}{\hat{Na}_{i,j}}^\top
\end{aligned}
\label{equation8}
\end{equation}
Where,$AttnMat_{i,j}$ is one element of $AttnMat$ with rich historical information. Further, the attention matrix is multiplied with $V_{c}$, and the value of each point in $f_{c}$ is modified by combining the Spatio-temporal information.

Temporal information can enhance the template feature, as it contains rich historical information and prior knowledge. Unfortunately, a priori misleading may occur. In particular, the state change of the target in two consecutive frames is too large. To augment the reliability of the final template feature, we use the convolution layer as the adaptive filter $\Phi_{MF}$ to correct the prior misleading. It is worth noting that the correction ability of adaptive filtering is limited. If continuous extreme conditions occur, the ST's output will lose reliable feature information. Consideration from two aspects of gradient back-propagation and feature augmentation, we continue to add template feature $f_{i}$ of the initial ground-truth frame to the output after the $\Phi_{MF}$ correction to get the enhanced template feature $f^{*}_{z}$. This ensures that the real target information will not be lost while integrating the Spatio-temporal information so that our tracker can better use the temporal information to enhance the robustness of the tracker.\\
\subsection{Discriminative augmentation module(DA)}
\quad After getting $f^{*}_{z}$ through the improved encoder, to improve the ability to distinguish targets in the search region $T_{s}\in \mathbb{R}^{C\times N_{s}\times N_{s}}$, we use DA module to augment the search region feature $f_{s}\in \mathbb{R}^{C\times \dfrac{N_{s}}{stride}\times \dfrac{N_{s}}{stride}}$. DA can be formulated as follows:
\begin{equation}
DA(f^{*}_{z},f_{s})=\Phi_{DA}(Decoder(f^{*}_{z},f_{s}))+f_{s}
\end{equation}
The improved decoder is proposed to augment $f_{s}$ before performing cross-correlation operations. Firstly, use the self-attention mechanism to make the feature points in $T_{s}$ pay more attention to other similar feature points. Then use the cross attention mechanism to integrate $f_{s}$ and $f^{*}_{z}$. Finally, a discriminative mask is generated. The mask's size is consistent with $f_{s}$. As with the template, we use convolution layers $\Phi_{DA}$ to suppress interference information of the mask. Finally, mask and $f_{s}$ make element-wise addition to get the enhanced search region feature $f^{*}_{s}$, which is used for subsequent cross-correlation calculation with $f^{*}_{z}$ to obtain the final reliable response map. The two proposed modules can be found in Figure~\ref{fig3}.

\subsection{Training and inference}
\begin{enumerate}
\item \textbf{Offline training.} First, randomly select 3 frames from one video sequence as different templates, that is $T_{i}$, $T_{a}$, $T_{c}$. After feature extraction of backbone, we get $f_{i}$,$f_{a}$,$f_{c}$. The range of selection is 50 consecutive frames in the video. For $T_{i}$, we directly randomly select frames from 50 consecutive frames. In the image augmentation stage of preprocessing, to improve the tracking system's robustness against corroded frames, noise information is randomly added to $T_{a}$, and $T_{c}$ respectively.When selecting frames, two consecutive frames will be randomly selected as $T_{a}$ and $T_{c}$ to ensure the module's ability to integrate temporal information. To fit the inference stage execution process to the great extent.
\item \textbf{Loss function.} For regression, we use smooth-$L_{1}$ loss consistent with SiamRPN to predict the normalized distance from the anchor center to the ground-truth center. For classification, we use two branches. One is the same as SiamRPN, which employs cross-entropy loss to predict positive and negative samples in the predicted response map. we add another classified branch using binary cross entropy(BCE) concerning FCOS \cite{tian2019fcos} to augment the confidence for adaptive template update. In SiamRPN, the classification branch adopts the IoU (Intersection over Union) based division method for positive and negative samples. For each anchor, compute their IoU values. If IoU values are greater than the presetting threshold, set these anchor boxes as positive samples, otherwise set them as negative samples. According to the analysis in \cite{zhang2020bridging}, the method of label assignment based on center distance can bring higher mAP than that based on IoU, and the regression method has little effect on the results of mAP. In our methods, we choose to use the scheme based on the distance from the anchor boxes' center point to the ground-truth center point on classified branches, which can be formulated as follows:
\begin{equation}
\begin{aligned}
&c_{pos}=\phi(C_{pos})=\dfrac{C_{pos}-ori}{stride}\\
&distance_{x_{i},y_{i}}=(\dfrac{c_{y_{lt}}+c_{y_{rb}}}{h}-row_{i})^{2}\\
&\qquad\qquad\quad\quad+(\dfrac{c_{x_{lt}}+c_{x_{rb}}}{w}-col_{i})^{2}
\end{aligned}
\end{equation}
$C_{pos}$ is the corner coordinate of bounding box, where $pos$ includes left top($lt$) coordinate and right bottom($rb$) coordinate. With function $\phi$, we can transform the original coordinate values of the bounding box into the response feature space of classified branches' output and $ori$ is presetting the origin value of the anchor. After obtaining $c_{pos}$, we can compute the distance from the center point of the mapped bounding box to the center point of each anchor on the feature map, that is $distance_{x_{i},y_{i}}$.$row_{i}$ represents the ith row of feature map and $col_{i}$ represents i-th col of the feature map. Further set anchor boxes whose $distance_{x_{i},y_{i}}$ is less than the threshold value to positive samples, otherwise set them to negative samples.

Final total loss $L_{total}$ function is:
\begin{equation}
\begin{aligned}
&L_{cls}=\lambda L_{cls1}+L_{cls2}\\
&L_{total}=\lambda_{1} L_{cls}+L_{reg}
\end{aligned}
\end{equation}
where $\lambda$ and $\lambda_{1}$ are hyperparameters to balance.
\item \textbf{Inference.} $f_{i}$ represents the initial frame, which is the most reliable reference template and will never change in the whole tracking process, $f_{c}$ is the current template updated after tracker processes each frame, and $f_{a}$ is the result of ST's output. Initially use the feature map of the initial frame $T_{i}$ as the $f_{i}$, $f_{a}$, $f_{c}$, where $f_{i}$ is framed in the initial ground-truth frame. In each subsequent frame, we use a weighted sum of the prediction results of the classified branches based on the center distance to select the target with the highest confidence and send it to the ST to enhance our template in the next step.
\end{enumerate}

\newcommand{\ie}{\textit{i}.\textit{e}.}

\section{Experiments}

\subsection{Implementation details}  
\begin{enumerate}
\item \textbf{Method detail.} We use SiamFC and SiamRPN as our base trackers. We have not made any changes to the original version of SiamFC. For SiamRPN, we change its backbone to modified ResNet50\cite{he2016deep}. Simultaneously modify the label assignment of SiamRPN by changing the anchor setting based on IoU to the setting based on the center distance. It simplifies the setting of the pipeline and hyperparameters and reduces the number of parameters. For the proposed module embedding scheme, different embedding methods are used according to the differences between SiamFC and SiamRPN. Due to the existence of feature pyramids in SiamFC, the discriminative augmentation module does not involve the processing of multi-level features, only spatio-temporal fusion encoders can be embedded. In SiamRPN, both two modules can be embedded. In SiamRPN, through many experiments, we decide to set the update threshold to 1.18, which will minimize the chance of template deterioration. In SiamFC, similar to the modified SiamRPN, the threshold value is directly set for the maximum value of the response map. For convenience, we use DASTSiam to refer to SiamRPN modified by our methods.
\item \textbf{Training detail.} The training data consists of the train-splits of LaSOT\cite{fan2019lasot}, VID\cite{russakovsky2015imagenet}. DASTSiam is trained over the course of 50 epochs. The backbone's parameters are fixed for the first 10 epochs. The learning rate in log space decreases from 0.005 to 0.0005 for the remainder of the training procedure. With a momentum of 0.9, SGD is used as the optimizer, and the mini-batch size is 12 pairings. Templates and search pictures are 287*287 pixels and 127*127 pixels, respectively, in size.
\item \textbf{Platform.} Our trackers are implemented using Python 3.6 and PyTorch 1.1.0.The experiments are conducted on a server with NVIDIA 12G 3080Ti GPU.
\item \textbf{Benchmark.} We evaluate results on several public tracking benchmarks:VOT2018 \cite{kristan2018sixth}, LaSOT test set\cite{fan2019lasot}, OTB100 \cite{wu2013online}, GOT-10k \cite{huang2019got}.
\end{enumerate}

\subsection{Comparisons}
To fully measure our approach. The above four benchmarks are used to evaluate the performance of our method and compare it with other state of art tracking algorithms.
\begin{figure*}[t!]
\centering
\subfloat[\label{fig4:a} OTB100 success plots]{
	\includegraphics[width=0.42\textwidth,height=0.35\textwidth]{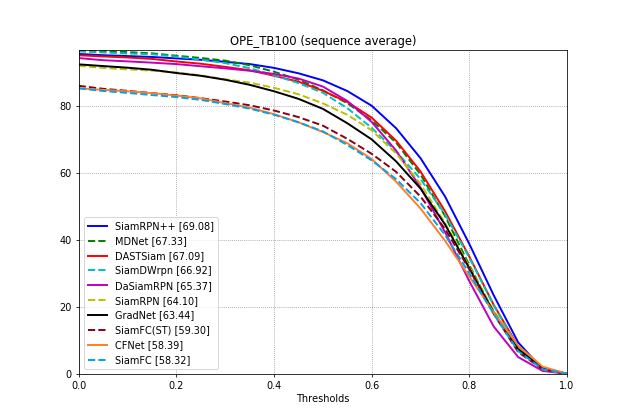}}
\hspace{16mm}
\subfloat[\label{fig4:b} OTB100 precision plots]{
	\includegraphics[width=0.42\textwidth,height=0.35\textwidth]{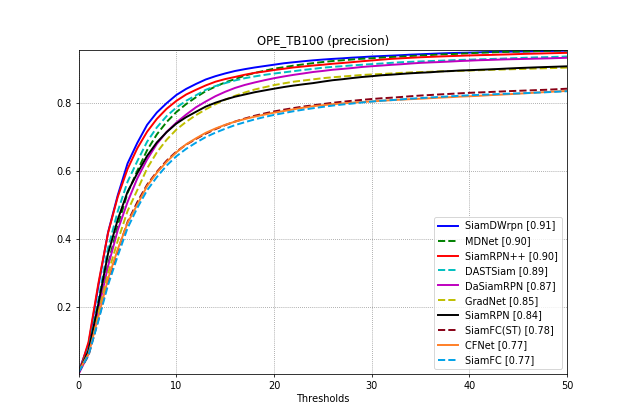}}
\caption{Success and precision plots of OPE show a comparison of our tracker with state-of-the-art trackers on the OTB100 dataset.}
\label{fig4}
\end{figure*}
\begin{figure*} [t!]
	\centering
	\subfloat{
		\includegraphics[scale=0.25]{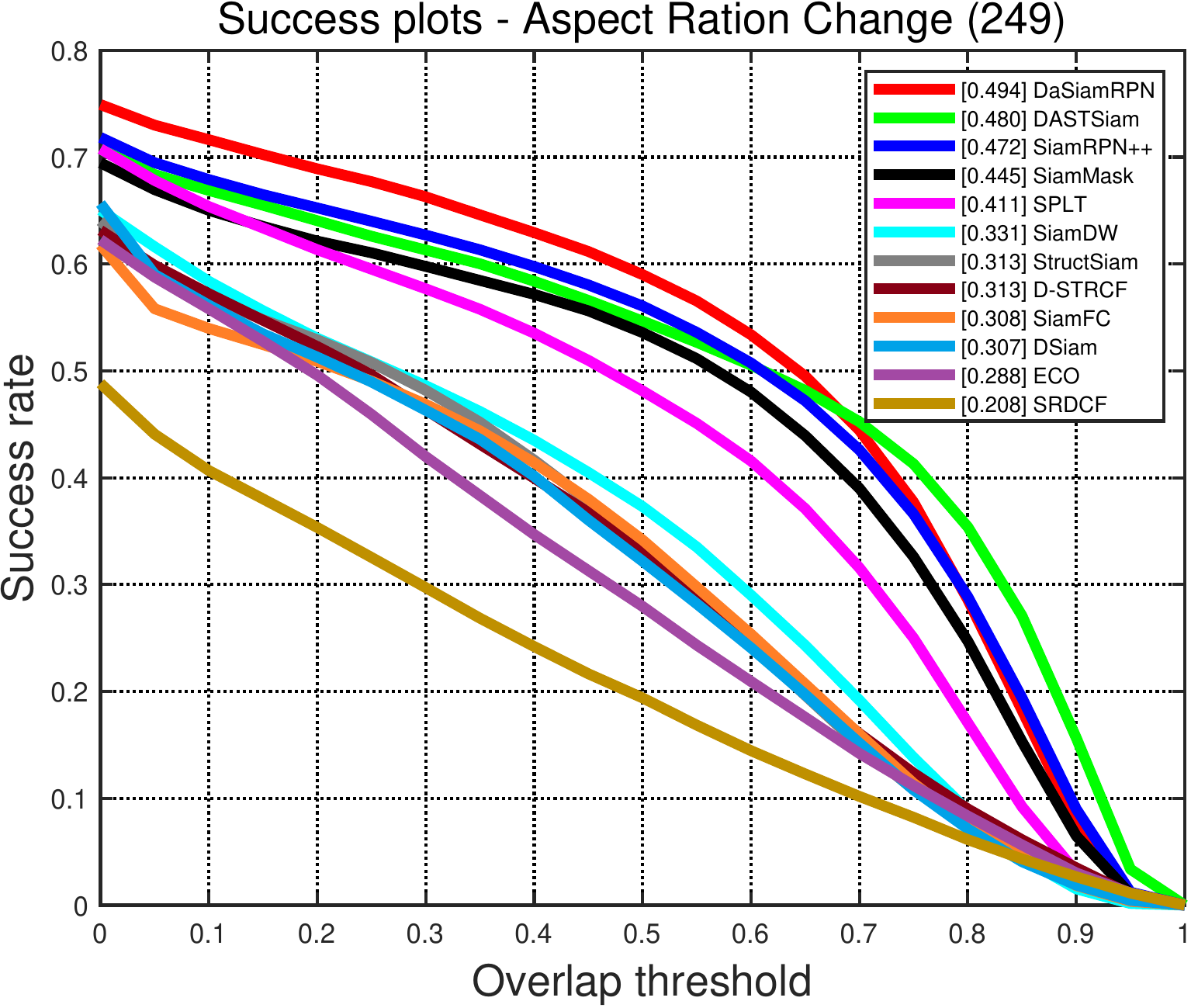}}
	\subfloat{
		\includegraphics[scale=0.25]{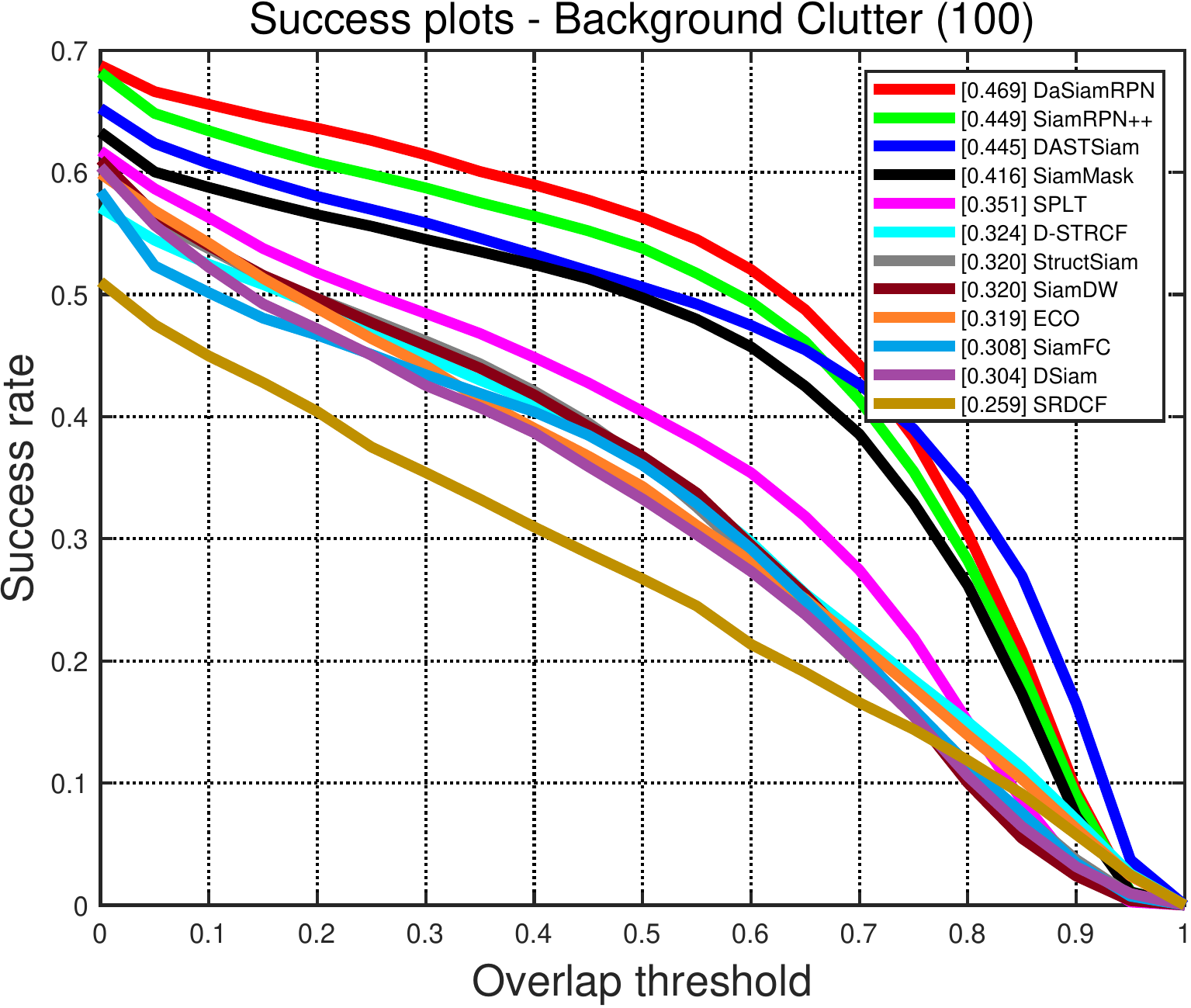}}
	\subfloat{
		\includegraphics[scale=0.25]{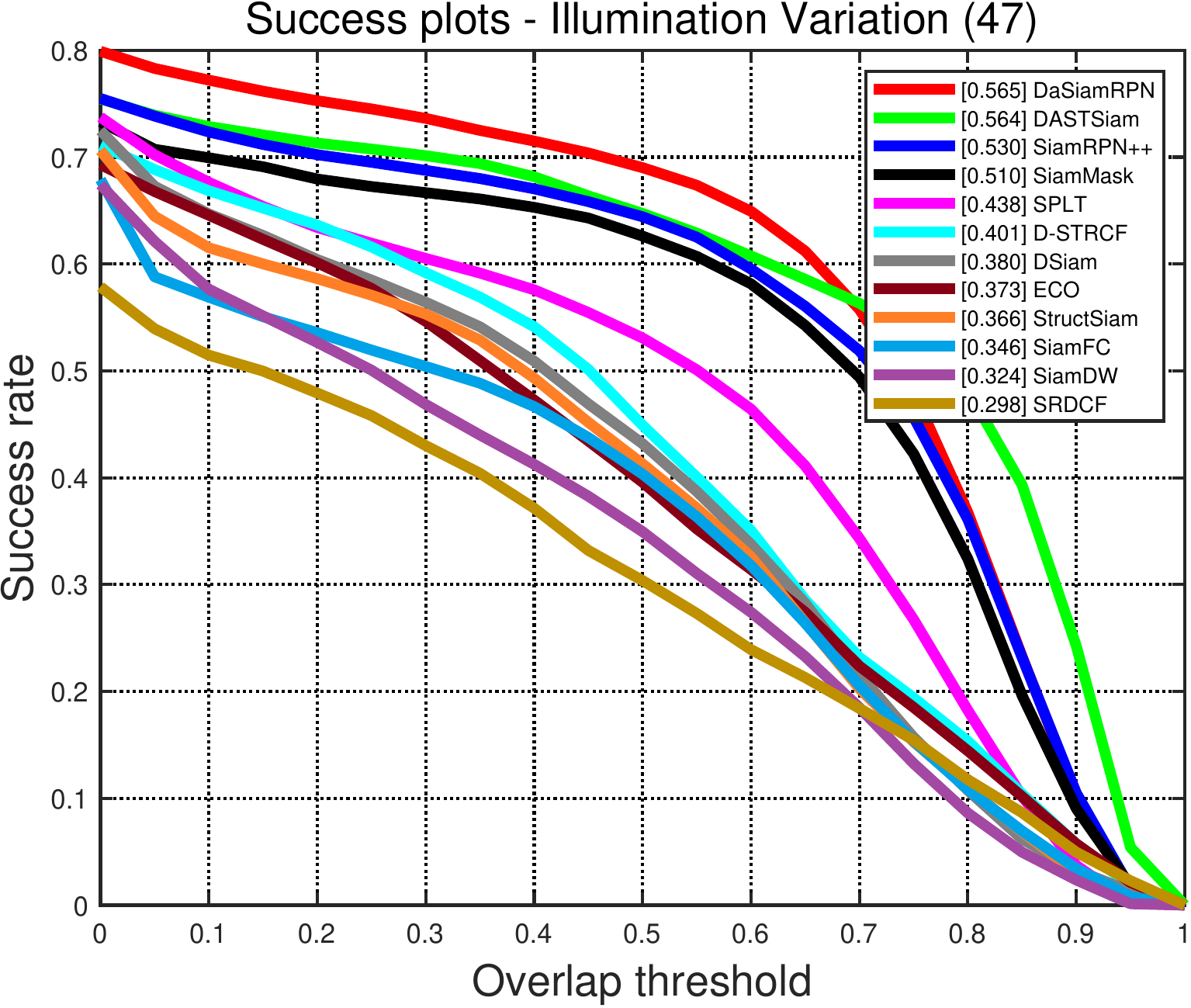}}
	\\
	\subfloat{
		\includegraphics[scale=0.25]{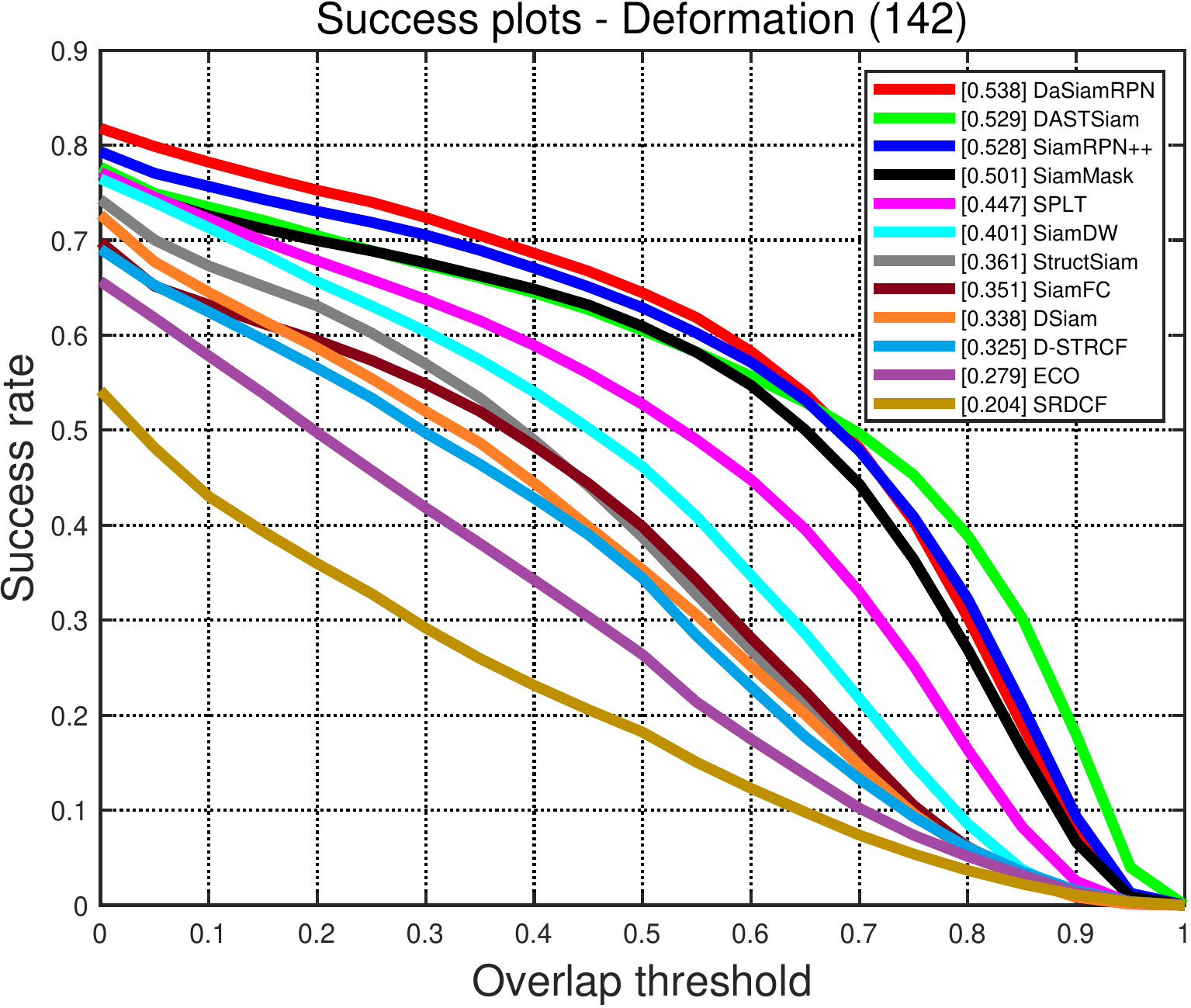} }
	\subfloat{
		\includegraphics[scale=0.25]{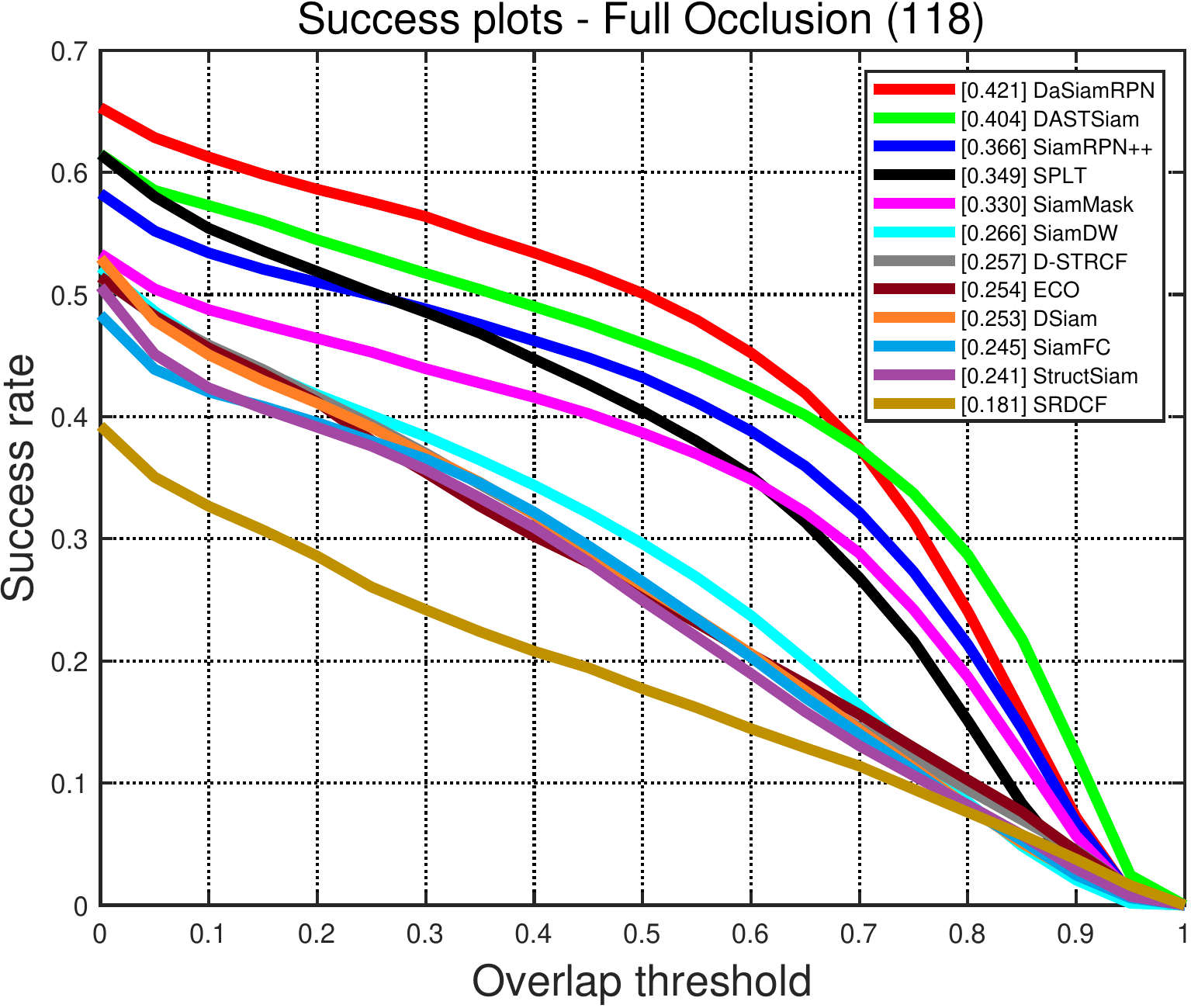}}
	\subfloat{
		\includegraphics[scale=0.25]{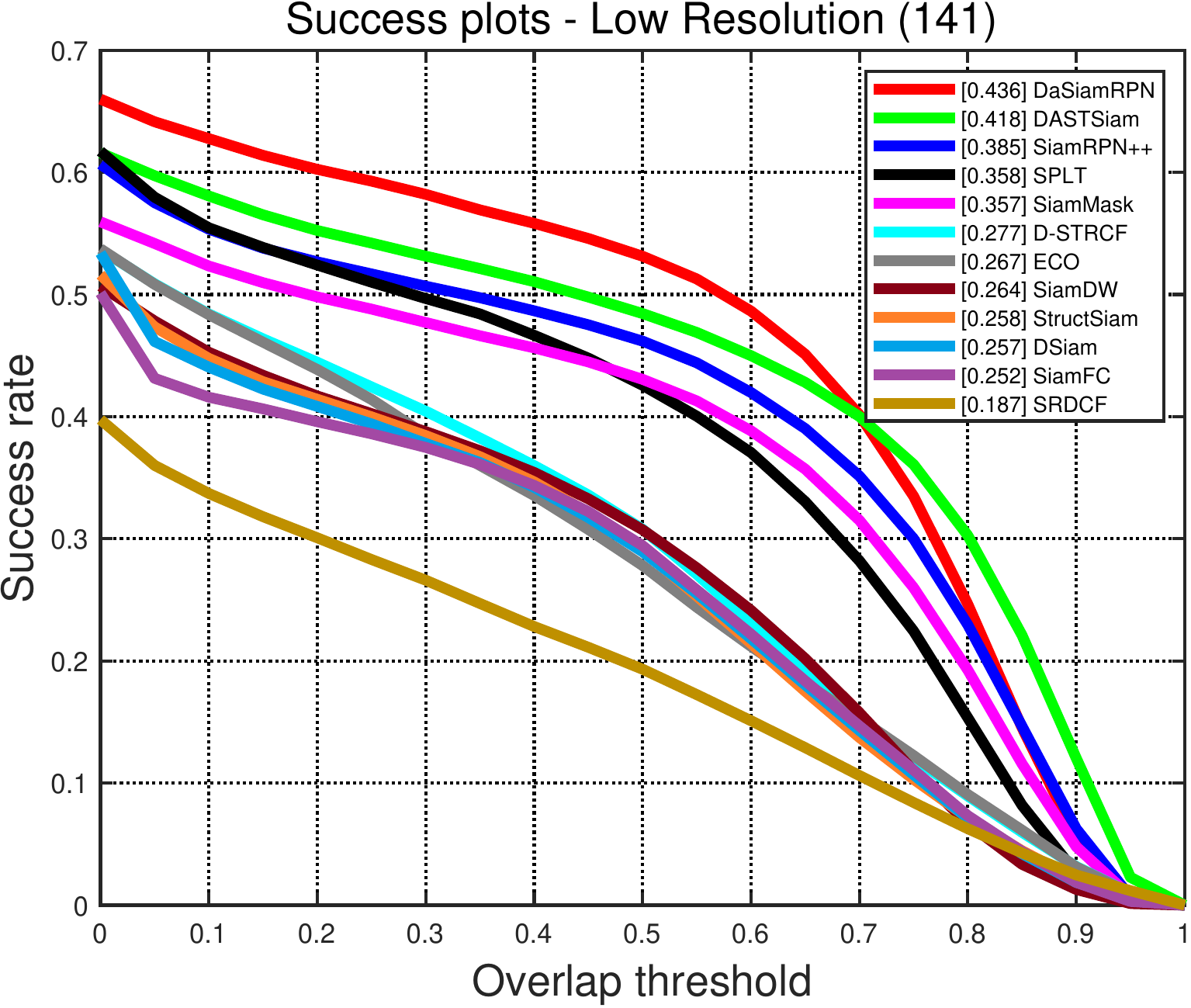}}
	\\
	\subfloat{
		\includegraphics[scale=0.25]{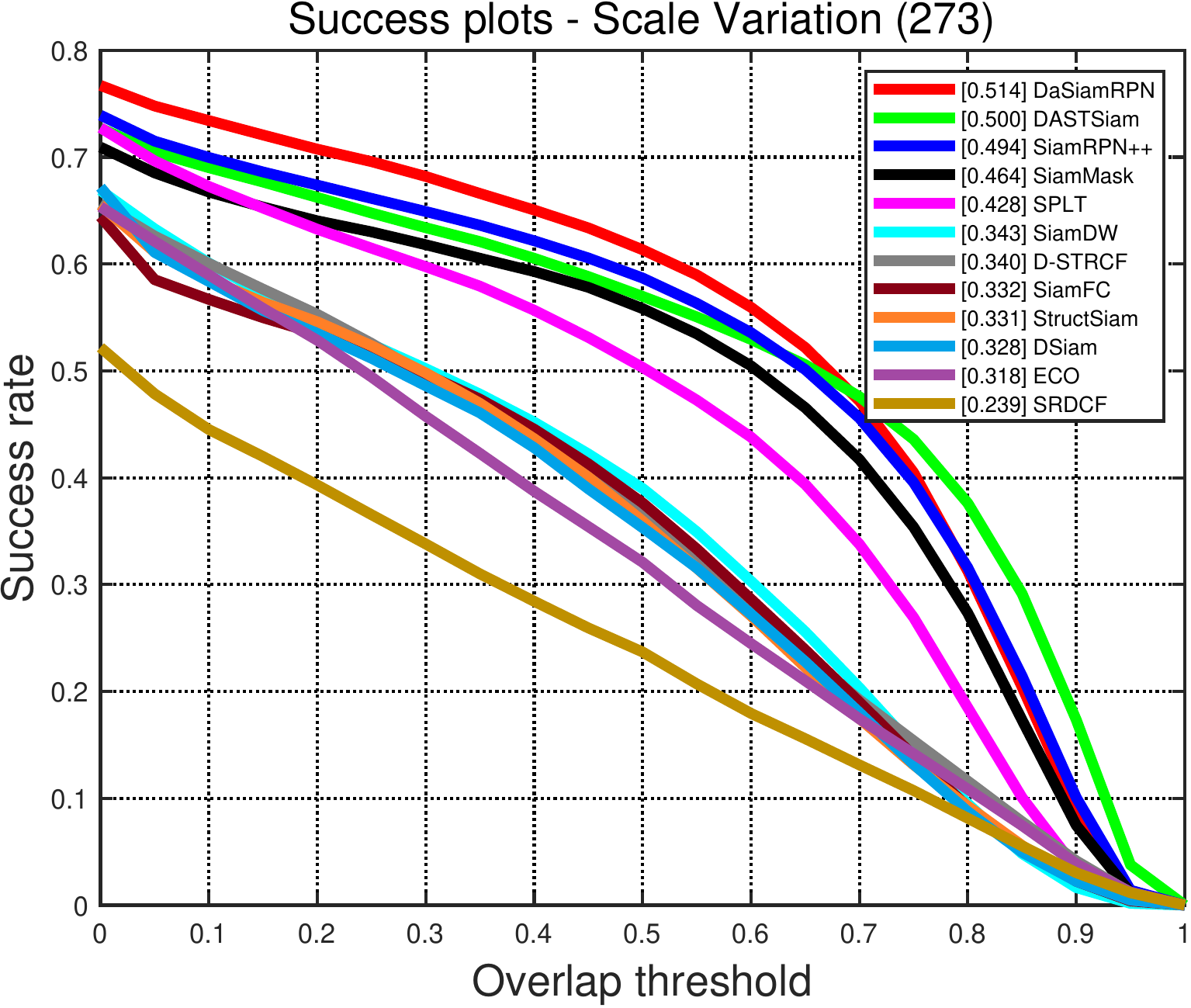} }
	\subfloat{
		\includegraphics[scale=0.25]{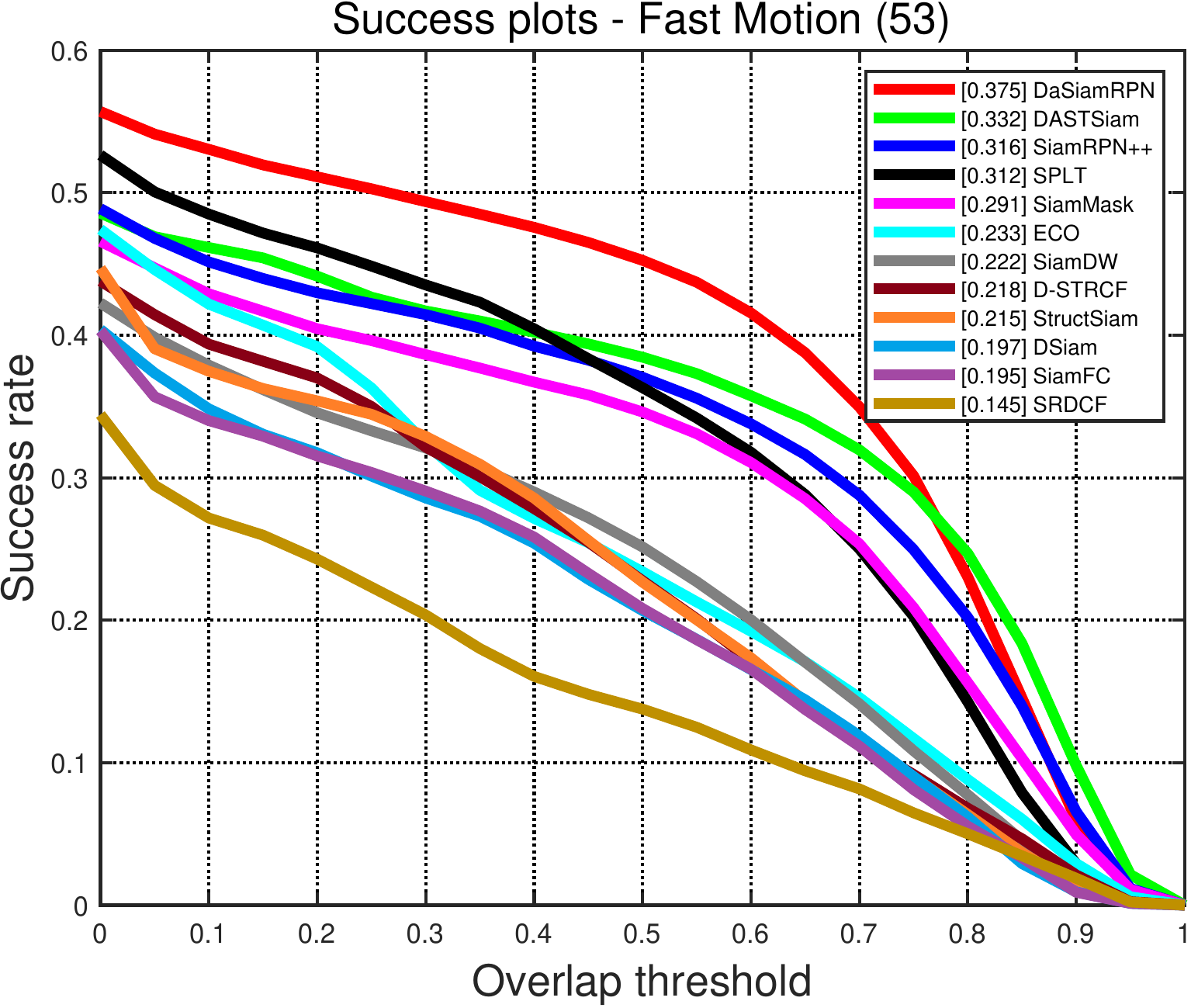}}
	\subfloat{
		\includegraphics[scale=0.25]{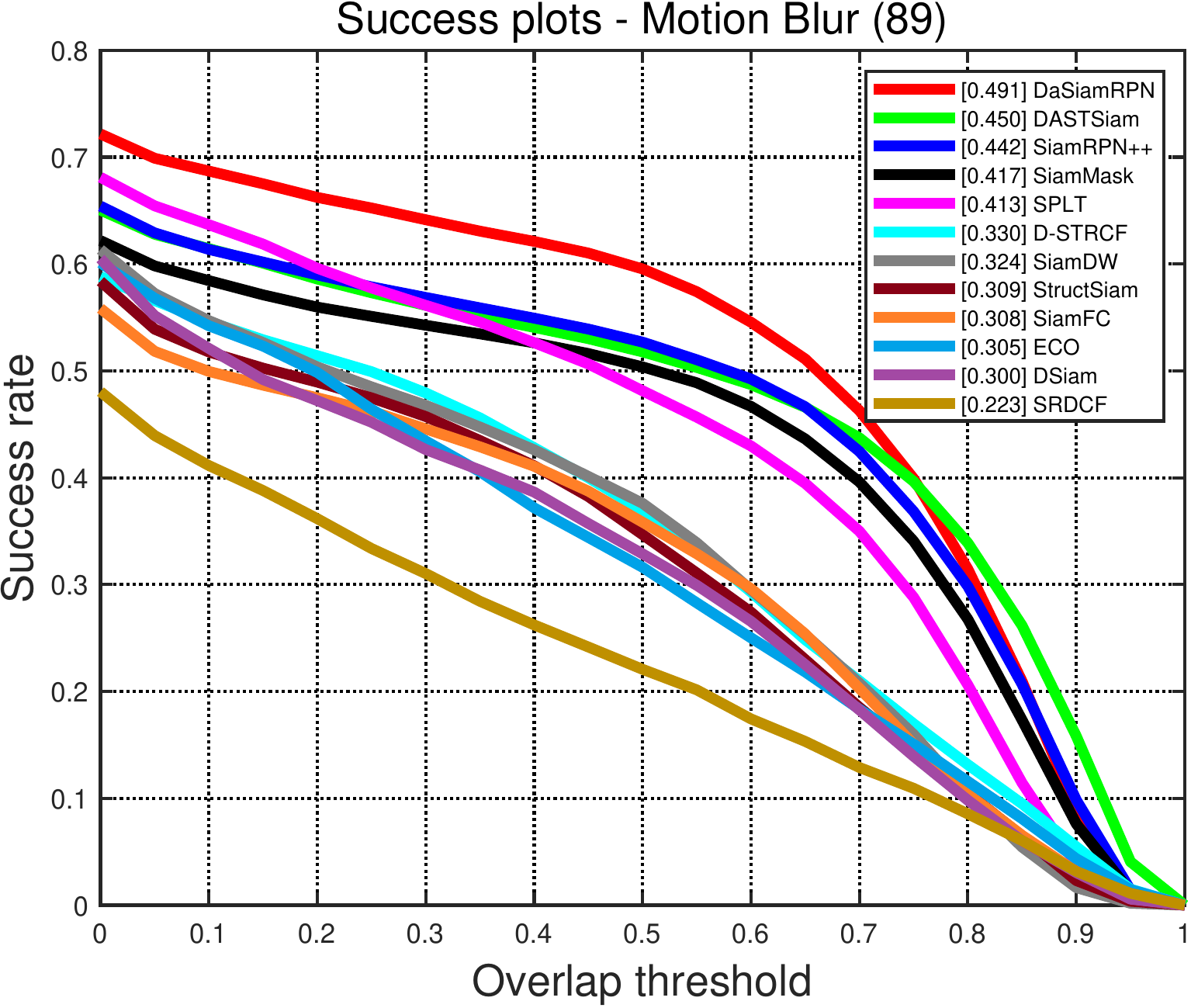}}
	\caption{ Attribute-based evaluation. The experimental results on 9 different challenging factors (including aspect ratio change, background clutter, illumination variation, deformation, full occlusion, low resolution, scale variation, fast motion, and motion blur.) are presented. The number of video sequences for each attribute challenging factor is shown in the parenthesis.}
	\label{fig5} 
\end{figure*}
\begin{figure*} [t!]
	\centering
	\subfloat[\label{fig6:a} LaSOT success plots]{
		\includegraphics[width=0.4\textwidth,height=0.28\textwidth]{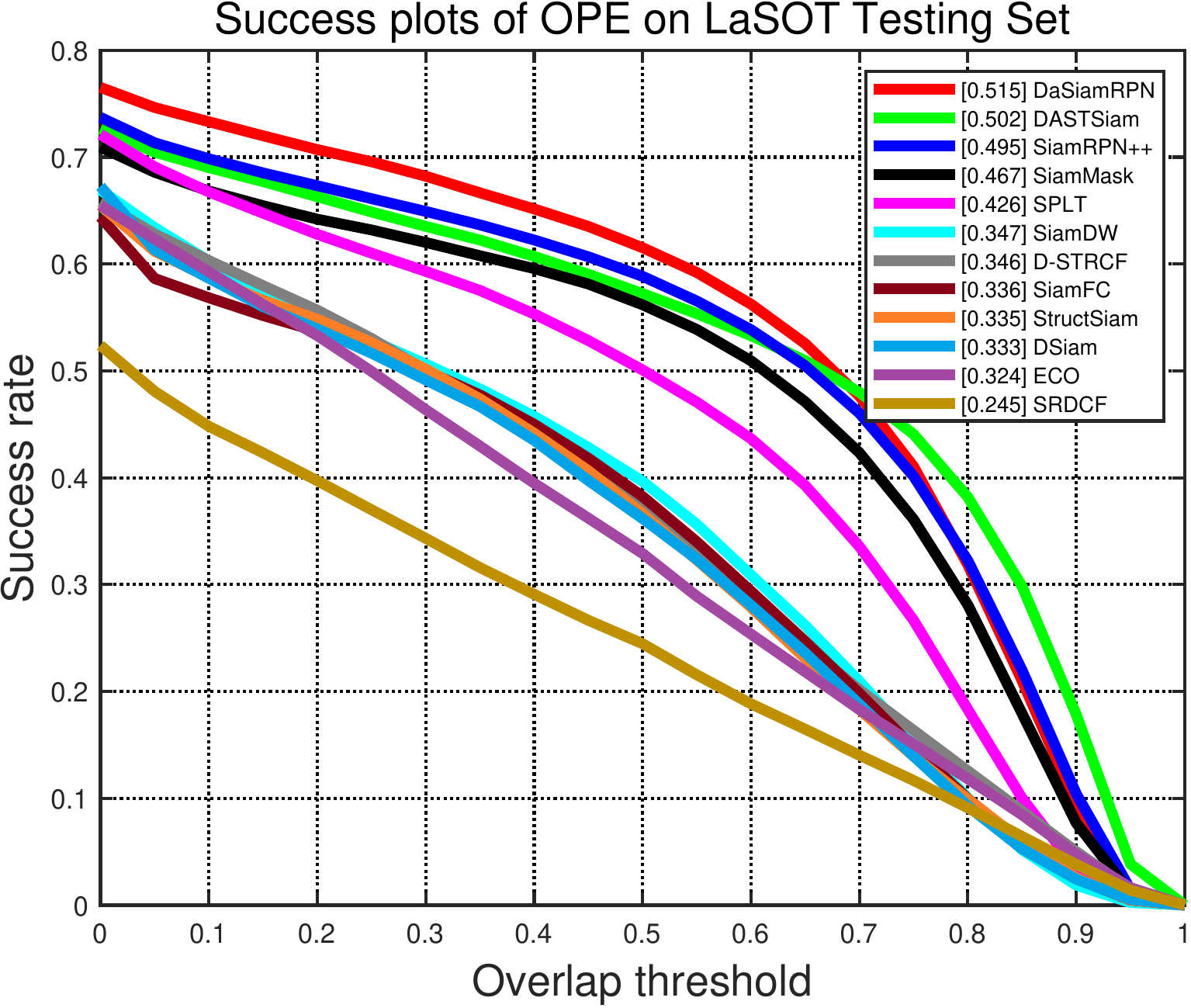}}
	\subfloat[\label{fig6:b} LaSOT precision plots]{
			\includegraphics[width=0.4\textwidth,height=0.28\textwidth]{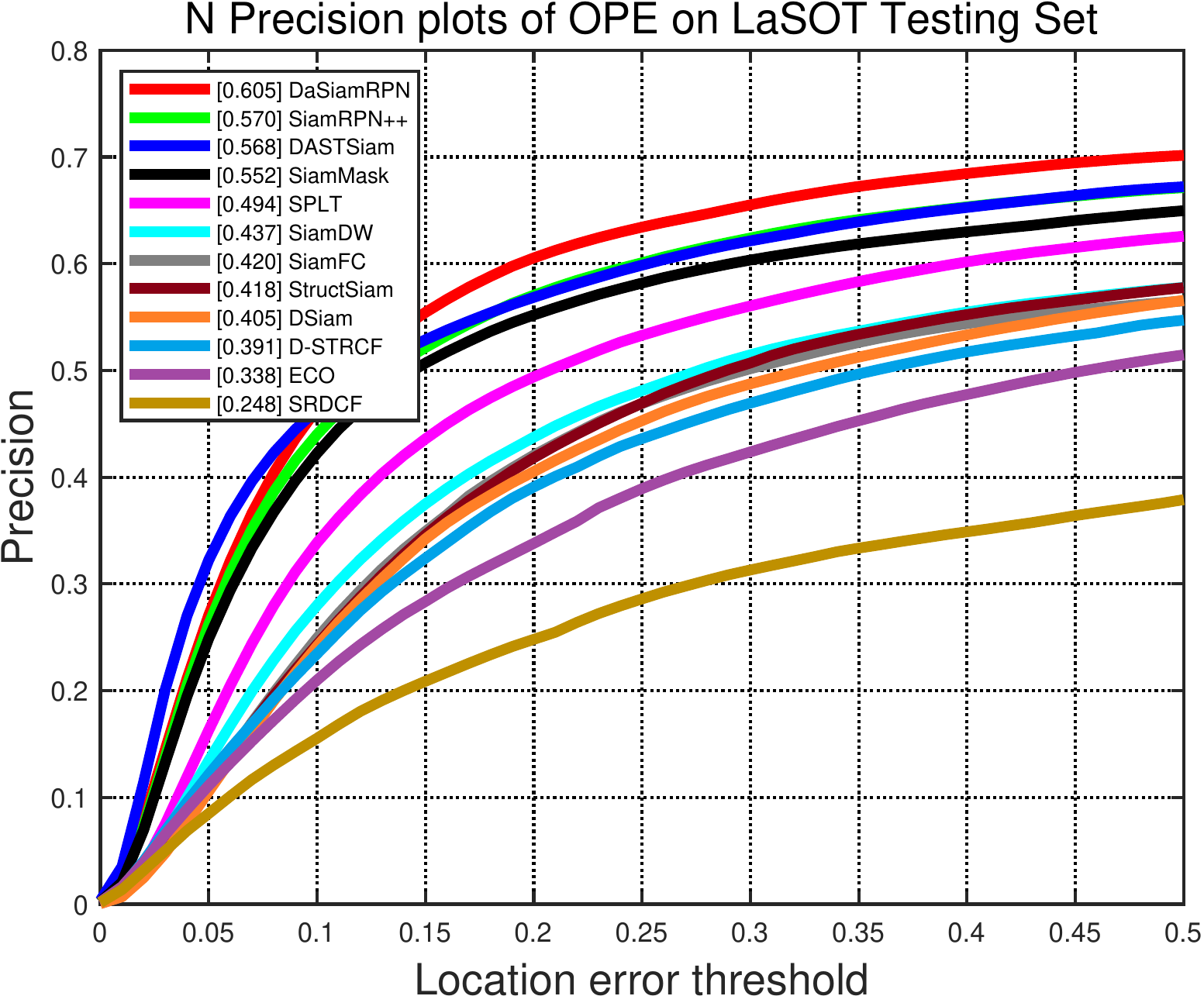}}
	\caption{ The precision and success plots of OPE on the LaSOT dataset. The solid lines in different colors represent the results obtained by 11 different tracking methods. The center location errors for precision plots and the average overlap scores for success plots are respectively shown in the legend. (a) LaSOT success plots. (b) LaSOT precision plots. }
	\label{fig6} 
\end{figure*}

First, we use our proposed methods and several state of art algorithms to make testing and evaluating on OTB100. DASTSiam is based on SiamRPN with spatio-temporal fusion and discriminative augmentation modules. The experimental results show that the tracking performance of the two modules has been improved by embedding them in the baseline. SiamFC's success improved from 58.32 to 59.30. SiamRPN's success is increased from 64.10 to 67.09. 
\begin{figure} [t!]
	\centering
	\subfloat{
		\includegraphics[width=0.43\textwidth,height=0.35\textwidth]{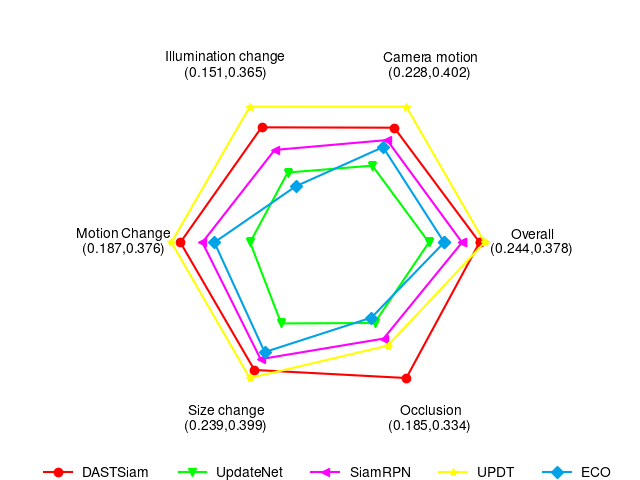}}
	\caption{Comparison of EAO on VOT2018 for the following visual attributes: camera motion, illumination change, occlusion, size change, and motion change. The values in parentheses indicate the EAO range of each attribute and overall of the trackers}
	\label{fig7} 
\end{figure}
In comparison, DASTSiam carries out a complete module embedding, it enhances the temporal information integration and discrimination ability of the tracker. At the same time, the label assignment based on the target center distance is adopted. Finally, the confidence score is calculated by multi-branch weighting, which improves the confidence of the adaptive template update. Thus, DASTSiam has been greatly improved based on the baseline. As shown in Figure \ref{fig4}, it can be found that DASTSiam has improved in both the success and precision indicators, and its performance exceeds the state of art algorithms such as CFNet, GradNet, SiamDW, and DaSiamRPN.
\begin{figure} [t!]
	\centering
	\subfloat{
		\includegraphics[width=0.42\textwidth,height=0.40\textwidth]{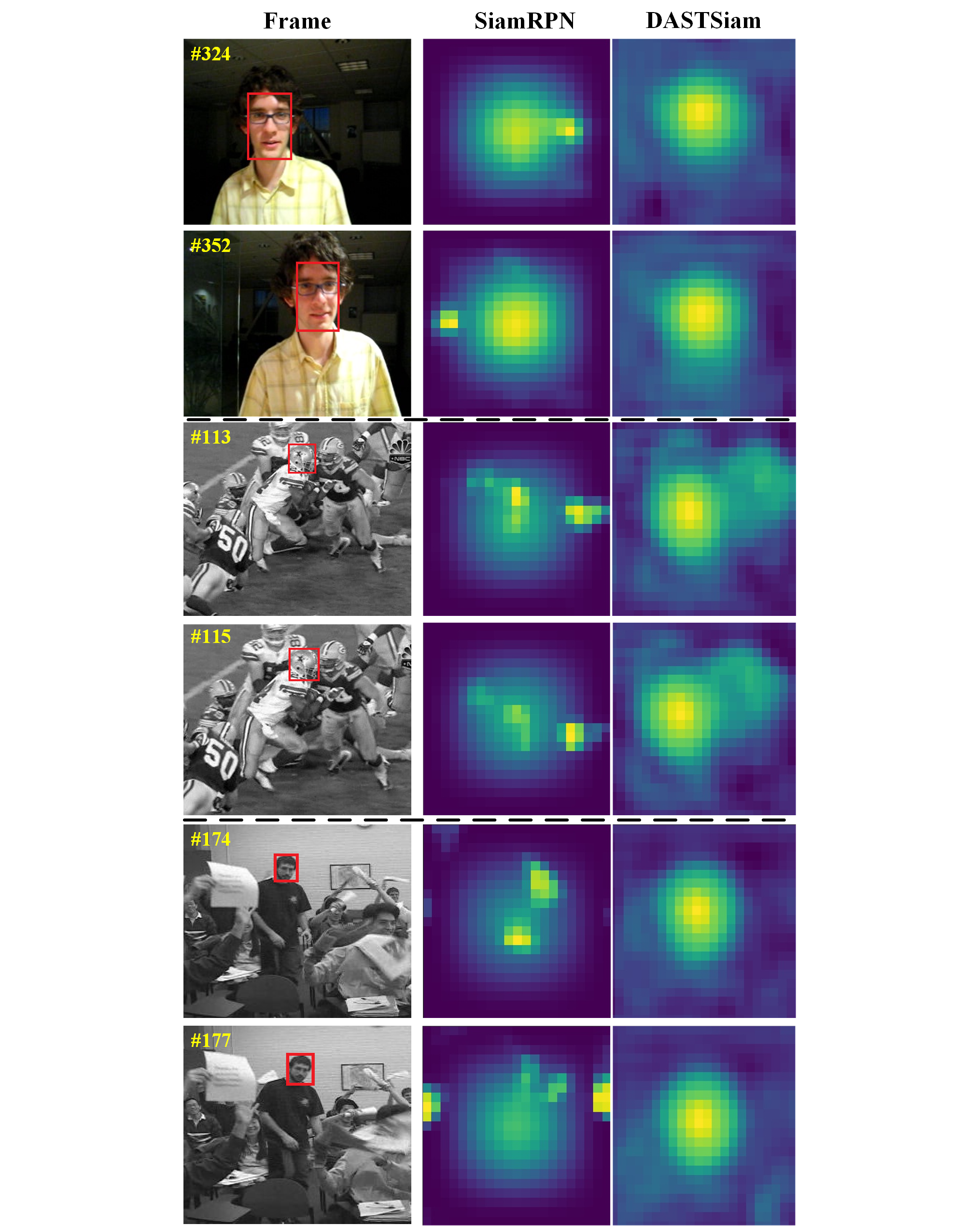}}
	\caption{Visualization of Classified response maps for SiamRPN and DASTSiam. 'Frame' column shows the search region and ground-truth box used to extract the templates. 'SiamRPN' presents the response map resulting from applying the initial template to the search region in SiamRPN. For 'DASTSiam', use a spatio-temporal fusion encoder module and discriminative augmentation decoder module to enhance our features of template and search region to get more accurate classified response maps.}
	\label{fig8} 
\end{figure}
\begin{table}[htbp]%调节图片位置，h：浮动；t：顶部；b:底部；p：当前位置
	\centering
	\renewcommand{\arraystretch}{0.8}
    \setlength{\tabcolsep}{5pt}
	\caption{Comparisons on VOT2018(\textcolor{red}{red} represents the best value, and \textcolor{green}{green} represents the second value.)}
	\label{table1}
	\scalebox{0.85}{
	\begin{tabular}{c|ccccc}
		\hline\hline	
		Trackers & UpdateNet & UPDT & SiamRPN& ECO& DASTSiam  \\
	    \hline
		Accuracy &0.518 &0.536 &\textcolor{green}{0.576} &0.484 &\textcolor{red}{0.585} \\
		Robustness & \textcolor{red}{0.454} & 0.184 &\textcolor{green}{0.323} &0.276 &0.295 \\
		EAO &0.244 &\textcolor{red}{0.378} &0.324 &0.280 &\textcolor{green}{0.366} \\
		\hline
	\end{tabular}
	}
\end{table}
\begin{table}[htbp]%调节图片位置，h：浮动；t：顶部；b:底部；p：当前位置
	\centering
	\renewcommand{\arraystretch}{0.8}
    \setlength{\tabcolsep}{5pt}
	\caption{Comparisons on GOT-10k(\textcolor{red}{red} represents the best value, and \textcolor{green}{green} represents the second value.)}
	\label{table2}
	\scalebox{0.85}{
	\begin{tabular}{c|ccccc}
		\hline\hline
		Trackers &ECO &SiamRPN++ &ATOM &SiamCAR &DASTSiam  \\ %\textcolor{red/blue/green/black/white/cyan/magenta/yellow}{text}
		\hline
		$AO$ &0.316 &0.517 &0.556 &\textcolor{red}{0.569} &\textcolor{green}{0.567} \\
		$SR_{50}$ &0.309 & 0.615 &0.634 &\textcolor{red}{0.670} &\textcolor{green}{0.656} \\
		$SR_{75}$ &0.111 &0.329 &0.402 &\textcolor{green}{0.415} &\textcolor{red}{0.452} \\
		\hline
	\end{tabular}
	}
\end{table}

\begin{table}[!ht]
	
	\centering
	\renewcommand{\arraystretch}{0.8}
	\setlength{\tabcolsep}{5pt}
	\caption{ablation study for verification of update strategy }
	\label{table3}
	\begin{threeparttable}
		\resizebox{\linewidth}{!}{
			\begin{tabular}{c|cc|ccc}\hline\hline\noalign{\smallskip}
				\multicolumn{1}{c|}{\multirow{2}{*}{label assignment}} & \multicolumn{2}{c|}{OTB100} & \multicolumn{3}{c}{GOT-10k} \\ \cline{2-6}
				&Success &Precision &$AO$&$SR_{50}$&$SR_{75}$ \\ \hline
				Anchor-IoU &59.86 &0.78 &0.416	&0.422 &0.153 \\ \hline
				Anchor-CenterDistance &\textbf{60.70} &\textbf{0.80} &\textbf{0.423} &\textbf{0.426} &\textbf{0.158} \\ \hline
			\end{tabular}
		}	
	\end{threeparttable}
\end{table}

\begin{table}[!ht]
	
	\centering
	\renewcommand{\arraystretch}{0.2}
	\setlength{\tabcolsep}{5pt}%30
	\caption{ablation study for verification of spatio-temporal fusion module(ST) }
	\label{table4}
	\begin{threeparttable}
		\resizebox{\linewidth}{!}{
			\begin{tabular}{c|c|cc}\hline\hline\noalign{\smallskip}
				\multicolumn{1}{c|}{\multirow{2}{*}{Trackers}} &\multicolumn{1}{c|}{\multirow{2}{*}{ST}} & \multicolumn{2}{c}{OTB100}\\ \cline{3-4}
				\multirow{2}{*}{ }&\multirow{2}{*}{ }&Success &Precision\\ \hline
				\multirow{2}{*}{SiamFC} & &58.32 &0.77\\ \cline{2-4}
				&$\checkmark$ &\textbf{59.30} &\textbf{0.78}\\ \hline
				\multirow{2}{*}{Modified SiamRPN} & &60.70 &0.80\\ \cline{2-4}
				&$\checkmark$ &\textbf{63.34} &\textbf{0.83}\\ \hline
			\end{tabular}
		}	
	\end{threeparttable}
\end{table}
\begin{table}[!ht]
	
	\centering
	\renewcommand{\arraystretch}{0.8}%1
	\setlength{\tabcolsep}{5pt}%10pt
	\caption{ablation study for verification of Discriminative augmentation }
	\label{table5}
	\begin{threeparttable}
		\resizebox{\linewidth}{!}{
			\begin{tabular}{c|c|c|cc|cccc}\hline\hline\noalign{\smallskip}
				\multicolumn{1}{c|}{\multirow{2}{*}{Trackers}} &\multicolumn{1}{c|}{\multirow{2}{*}{ST}} &\multicolumn{1}{c|}{\multirow{2}{*}{DA}} & \multicolumn{2}{c|}{OTB100}&\multicolumn{2}{c}{GOT-10k}\\ \cline{4-8}
				\multirow{2}{*}{ }&\multirow{2}{*}{ }&\multirow{2}{*}{ }&Success &Precision&$AO$&$SR_{50}$&$SR_{75}$\\ \hline
				\multirow{3}{*}{Modified SiamRPN} &  &  &60.70 &0.80 &0.423 &0.426 &0.158\\ \cline{2-8}
				&  &$\checkmark$ &\textbf{62.40} &\textbf{0.81} &\textbf{0.458} &\textbf{0.543} &\textbf{0.251}\\ \cline{2-8}
				&$\checkmark$ &  &63.34 &0.83 &0.463 &0.548 &0.254\\ \cline{2-8}
				&$\checkmark$ &$\checkmark$ &\textbf{67.09} &\textbf{0.89} &\textbf{0.567} &\textbf{0.656} &\textbf{0.452}\\
				\hline
			\end{tabular}
		}	
	\end{threeparttable}
\end{table}

To verify DASTSiam's ability to deal with typical tracking problems, we conducted a comprehensive performance test on the test set divided by the LaSOT data set. As shown in Figure \ref{fig5}. Compared with the baseline, DASTSiam has greatly improved in dealing with various tracking problems. In particular, to deal with deformation, aspect reason change, scale variation, and other issues, the spatio-temporal fusion module is embedded to integrate temporal information, which enhances the feature of the matching template, to improve DASTSiam's robustness against appearance changes. The discriminative augmentation module is introduced to enhance the search region feature, making DASTSiam more robust to the background divider problem than the baseline. The overall performance results given by the success plot and precision plot are shown in Figure \ref{fig6}, which shows that our DASTSiam exceeds that of SiamRPN++, SiamMask, SiamDW, and other state of art trackers.

Further, use the VOT2018 dataset to evaluate through different protocols. The tracking performance of DASTSiam is measured by EAO calculated by two indicators: Accuracy and robustness. As shown in Figure \ref{fig7}, in VOT2018, our method has better tracking performance than the baseline in terms of light change, camera motion, motion change, size change, and occlusion, refer to Table \ref{table1} for specific information. At the same time, we evaluated and verified DASTSiam on GOT-10k. The protocol proposed by GOT-10k, namely Class balanced metrics $AO$ and $SR$, is adopted to measure the tracking performance of our method. As shown in Table \ref{table2}, our method is superior to state of art algorithms such as SiamRPN++. It proves the effectiveness of our method.
\subsection{Ablation}
By choosing the same training datasets, train technique, and platform configurations, irrelevant interferences are reduced in order to fairly evaluate and validate the efficacy of proposed modules. The benchmark model for comparison are SiamFC and SiamRPN.

First, use the original SiamRPN and modified SiamRPN to verify the effectiveness of the template update scheme. Both adopt the same backbone ResNet50\cite{he2016deep}.  By comparing OTB100 and GOT-10k, As shown in Table \ref{table3}, we can find that the latter improves the performance of the tracker, which shows that it can bring more reliable template updates.

Second, To verify the spatio-temporal fusion module, we take modified SiamRPN and original SiamFC as the baseline. Keep the training strategy consistent .As shown in Table \ref{table4}, through experiments on OTB100, it is found that the performance of the tracker is improved when multi-frame fusion module is embedded into baselines the  respectively. The validity of the multi-frame fusion module is proved.

Finally, as shown in Table \ref{table5}, the discriminative augmentation module was verified by modified SiamRPN. First, we remove the spatio-temporal fusion module and then add the discriminative augmentation module. Testing on OTB100 and GOT-10K shows that the tracking performance is improved. 

\section{Conclusions}
In this paper, we proposed a novel approach for object tracking that utilizes the spatio-temporal fusion module (ST) and discriminative augmentation module (DA) to address the challenges of target deformation, occlusion, and scale variation in the tracking process. Our experimental results on four benchmark datasets (OTB100, laSOT, GOT-10k, VOT-2018) demonstrate the effectiveness of the proposed approach in improving the performance of the trackers.

In conclusion, our proposed method makes significant contributions to the field of object tracking by providing an effective solution for addressing the limitations of existing Siamese-based trackers in terms of exploiting spatio-temporal information and distinguishing targets from cluttered backgrounds. This work opens up opportunities for further research in areas such as feature adaptation and the application of the proposed method to other related tasks. We believe that our approach has the potential to significantly improve the performance of object tracking in real-world scenarios.

\section*{Acknowledgements}
Yucheng Huang and Eksan Firkat make an equal contribution. Thank Eksan Firkat, Jihong Zhu and Askar Hamdulla for their guidance.

\bibliographystyle{named}
\bibliography{ijcai23}

\end{document}